\documentclass[]{IEEEphot}

\usepackage[table,xcdraw]{xcolor}
\usepackage{float} 
\usepackage{amssymb}
\usepackage{amsmath}
\usepackage[]{graphicx}
\usepackage{rotating}
\begin{document}

\title{Image Inpainting using Partial Convolution}

\author{Amey Kulkarni, Harsh Patel,  
Shivam Sahni, Udit Vyas}

\affil{Indian Institute of Technology Gandhinagar, India}  


\maketitle

\markboth{Computer Vision Spring 2021}{Image Inpainting using Partial Convolution}


\begin{abstract}
Image Inpainting is one of the very popular tasks in the field of image processing with broad applications in computer vision. In various practical applications, images are often deteriorated by noise due to the presence of corrupted, lost or undesirable information. There have been various restoration techniques used in the past with both classical \cite{990497} \cite{Barcelos2007} and deep learning approach for handling such issues \cite{marinescu2021bayesian} . Some traditional methods include image restoration by filling gap pixels using the nearby known pixels or using the moving average over the same. The aim of this project is to perform image inpainting using robust deep learning methods that use partial convolution layers. 
\end{abstract}


\section{Problem Statement}

\textbf{Image Inpainting} is an image reconstruction technique, where missing sections of an image(holes) are filled or "predicted". This technique finds several applications, some of them include removing unwanted objects from an image, restore damaged portions of old images, or removing unwanted text. 
Classical vision approaches in Image Inpainting deal with the propagation of unmasked image parameters onto the holes. These methods have shown varied success, however, they all lack the ability to semantically reconstruct the holes.
Recent work on Image Inpainting has predominantly been based on learning-based approaches which can effectively learn the semantics of the image. Generative Adversarial Networks (GANs) have also been used in this domain to generate satisfactory results. \cite{marinescu2021bayesian}

This project is inspired from the original paper "Image Inpainting for Irregular Holes Using Partial Convolutions" \cite{DBLP:journals/corr/abs-1804-07723}. This approach of image inpainting is capable of restoring image with any arbitrary shaped holes. The central idea of this work involves in its Partial Convolution layers, and specific high-level feature loss networks. 

The upcoming section of the report contain sections describing the Approach, Quantitative and Qualitative results, and Key Observations. 

\section{Approach}
In this project, we make use of Partial Convolution Layers with mask update method to perform the inpainting operation. The following section describes the approach in detail. 

\begin{figure}
\centering
\includegraphics[width=0.5\textwidth, height=\textheight, keepaspectratio]{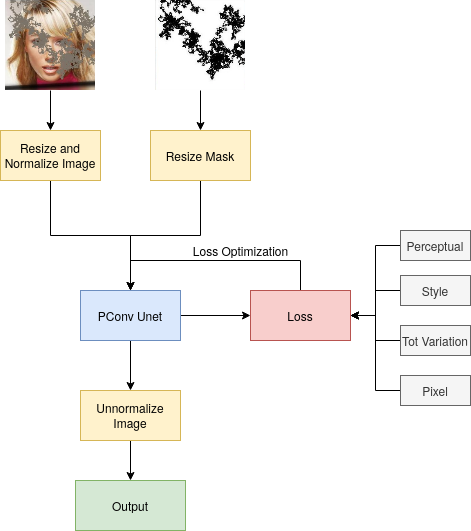}
\caption{Pipeline of the approach}
\label{fig:pipeline}
\end{figure}

\subsection{Dataset}
We use the CelebA-HQ/256 \cite{liu2015faceattributes}, a large-scale dataset that has 30,000 human face images each of size 256x256. We divided the dataset into train, validation and tests datasets in the proportion of 70\%, 15\%, and 15\%. For applications described in the later sections, we make use of samples of image size 256x256 from Places2 dataset \cite{zhou2017places}.

\subsection{Partial Convolution Layers}
A partial convolution, as the name suggests, is very similar to a usual Convolution layer except that it only performs the convolution operations on the pixels where the input pixel are not currently masked or "available". More formally, if \textbf{W} is the convolution filter, b is its corresponding filter, \textbf{M} is the binary mask corresponding to the convolution filter, and \textbf{X} are the image features, then the partial convolution is defined as- 

\begin{equation}
    x' = \begin{cases}
    W^{T}(X \circledast M)\frac{sum(\textbf{1}))}{sum(M)} + b, &  \text{ if } sum(M)>0 \\ 
    0, & \text{ if } otherwise 
    \end{cases}
\end{equation}
Note that \textbf{1} is a vector with same shape as M, containing all the elements as one.

\subsection{Mask Update Step}

Every Partial Convolution layer is followed by a mask update step. As seen in Section 2.2, only the \textit{available pixels} are used for calculating the convolution output. The mask update step is about updating a mask value from 0 to 1, if there is at least one valid input corresponding to the pixel under consideration. Mathematically, the updated mask $m'$ can be defined as -

\begin{equation}
    m' = \begin{cases}
    1, &  \text{ if } sum(M) \\
    0, & \text{ if } otherwise 
    \end{cases}
\end{equation}

It is important to observe that after successive iterations of the masked image through the network, all values of the mask will eventually be filled with ones, if the input contained at least one valid pixel.

\subsection{UNet Architecture}

\begin{figure}[!h]
\centering
\includegraphics[width=\textwidth, height=\textheight, keepaspectratio]{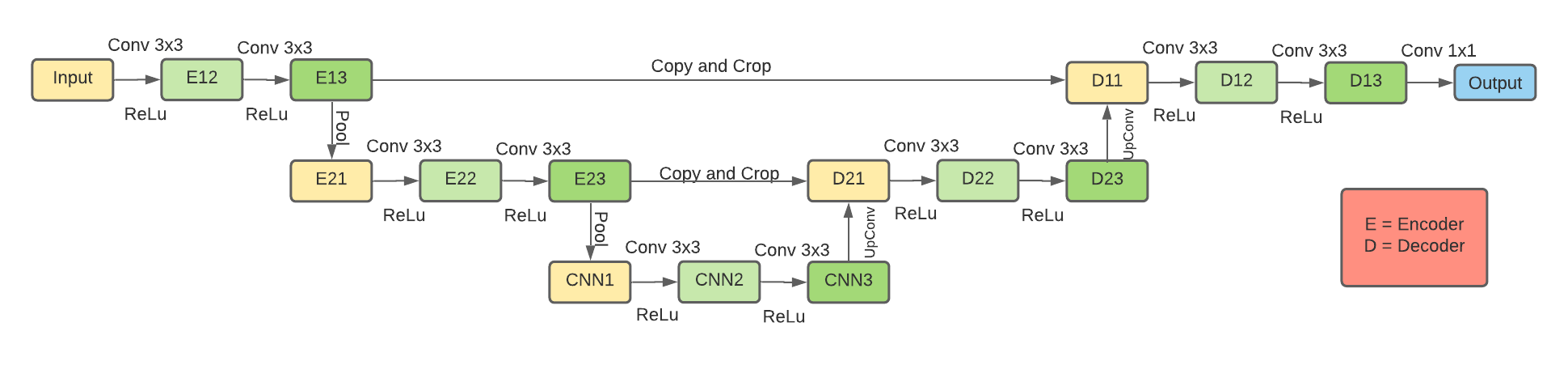}
\caption{Skeleton of typical UNET}
\label{fig:Unet}
\end{figure}

The UNet architecture, as can be seen from figure \ref{fig:Unet}, has a 'U' shape. In image \ref{fig:Unet} there are only two layers, however a typical UNet contains four. 

The two limbs of figure \ref{fig:Unet} represent the contraction(left portion, also called encoders) and expansion(right portion, also called decoders) paths respectively.

The contraction path is a stack of convolutional neural networks and max pooling layers applied successively. Every layer contains two CNNs, each followed by an activation function. Typically, the activation function for all layers is ReLu. The output of these two CNNs is passed through a max pooling layer with a kernel size that is typically equal to 2.
The convolution operations increase the number of channels(say from 3 for an RGB image) to a greater number of channels in a latent space(say 256 to 512). Since we apply max pooling(kernel size $\ge$2), the height and width reduce.

Functionally, the contraction path is used to capture information from images that helps us identify people, places or other entities relevant to us. In other words, it determines the "context" of the image. This can be understood by the fact that as the image goes through the contraction path, its size reduces, but the number of channels increase. This means that information regarding spatial locality is being lost, however information regarding the features in the image is being amplified.

The expansion path is also a stack of CNNs, but here instead of the max pooling layers, we have upsampling layers. Every decoding layer consists of two CNNs, each followed by an activation function(the last layer usually has no activation function).

An important property of UNet is the presence of skip connections. These refer to the connections that run across the pipeline (the "copy and crop" connections in image \ref{fig:Unet}). These separate it from regular encoder-decoder nets, that just use the output from the previous layer. The output from the corresponding layer(same depth) of the contraction path is concatenated with the output of the previous decoding layer.
Therefore the size of the image is gradually restored while the number of channels are reduced through the CNN layers.

The function of the expansion path is recovering the spatial locality information that was lost in the contraction path. Not only does this happen by upsampling, but also by the skip connections. This is because, the feature map from the encoder at the same depth level contains locality information.

All in all, the UNet architecture is one arranged in a U-shape, with a contraction path and an expansion path containing encoders and decoders respectively, and where shallow layers(encoders) are connected to deeper layers(decoders) through skip links.

\subsection{Modified UNet (Our Contribution)}
For our task of image inpainting, we have used a total of seven layers in our model. In other words, we have seven encoders and seven decoders. We have the same Module List inside each encoder and decoder as described in the previous section. However there are a few changes. 

We have used the following activation functions for different layers-
\begin{itemize}
    \item ReLu for all the encoding layers.
    \item Leaky ReLu for all decoding layers except the first layer(depth = 1).
    \item No activation for the first decoding layer.
\end{itemize}
Note that the first decoding layer is the topmost layer in the architecture and the final layer that is used in the sequence followed by the pipeline.

For the upsampling step we have used the bilinear mode of Pytorch since that is recommended for 4D data.

We have not required to crop the image from the encoder in any skip connection, since we have maintained the invariant that the image size at the $i^{th}$ encoder and decoder is $\frac{H}{2^(i-1)} \times \frac{W}{2^(i-1)}$. We ensure this by having padding of size $n$ for a kernel size of $2 \times n + 1$, with no stride or dilation for every convolution operation. This ensures that the output image size of any CNN is the same as its input image size.
The Max Pooling layer with kernel dimension of 2x2 is responsible for contraction by 2 while the Upsampling layer with scale factor = 2 is responsible for expansion.

\subsection{Loss Functions}
Choice of Loss functions plays a significant role in helping us to achieve both good predictions and faster convergence in any deep learning framework. In this image restoration problem, the following loss functions take charge in improving both features and their compositions in the restored image. For the subsequent sections please consider the following notation.
\begin{itemize}
    \item $I_{gt}$: Ground Truth Image
    \item $I_{out}$: Output Restored image generated by the network
    \item $I_{comp}$: Output Restored image with non-hole regions replaced with the ground truth.
    \item $\psi_{i}$: Activation map of the ith layer of the network
    \item $\xi(I)$ = Gram matrix of I
    \item $K_{h*w}$: Gram matrix normalising factor, dependent on the size of the input (h*w)
\end{itemize}

\subsubsection{Total Variation Loss}
The total variation loss is quite analogous to the nature of regularization losses. It is responsible for maintaining spatial continuity along with smoothness in the restored image. It is widely used in similar computer vision applications like Style Transfer and also, in digital signal processing for removal of noise.
  \begin{equation}
  \centering
    L_{tv} = \sum_{(i,j),(i,j+1)\epsilon I}^{} \left | I_{comp}^{[i,j+1]}\right | + \sum_{(i,j),(i+1,j)\epsilon I}^{} \left | I_{comp}^{[i+1,j]}\right |
    \end{equation}
\subsubsection{Perceptual Loss}
\label{p_loss}
Perceptual Loss also known as the feature reconstruction loss, encourages the pixels in the restored image to have similar feature representations rather than exactly matching the ground truth pixels \cite{johnson2016perceptual}. 

These feature representation are obtained by passing the image through a pre-trained Convolutional Neural Network (VGG-16 \cite{simonyan2015deep} in our experiments) which is a series of few convolution layers followed by some fully connected layers. The layers from the input to the last max pooling layer are used for the feature extraction.

This loss function tries to minimize the semantic differences between the features that are activated for the input image and for the reconstructed image, at the different available layers in the network. 
    \begin{equation}
        L_{perceptual} = \sum_{i = 0}^{N-1} \left | \psi_{i} (I_{out}) - \psi_{i} (I_{gt})\right | + \sum_{i = 0}^{N-1} \left | \psi_{i} (I_{comp}) - \psi_{i} (I_{gt})\right |
    \end{equation}
\subsubsection{Style Loss}

Style loss is used to incorporate texture like feature details of the ground truth image rather than its global arrangement into the restored image. It helps to capture general appearance of the ground truth image in terms of colors and localised compositions \cite{gatys2015neural}.

These style represented as gram matrices which are obtained by computing the correlations between the feature maps (same as in Section \ref{p_loss}) obtained from the activations of the convolutional neural network at different layers of the network.

This loss function tries to minimize the L1 distance between the entries of the Gram matrix from the style of the input image and both the Gram matrix of the reconstructed image and $I_{comp}$ generated after every iteration.

\begin{equation}
    \centering
    L_{style} = \sum_{i = 0}^{N-1} \left | \xi (\psi_{i} (I_{out})) - \xi(\psi_{i} (I_{gt}))\right | + \sum_{i = 0}^{N-1} \left | \xi (\psi_{i} (I_{comp})) - \xi(\psi_{i} (I_{gt}))\right |
\end{equation}
\begin{equation}
    \centering
    \xi (I_{h*w}) = K_{h*w}(I^{T}I) 
\end{equation}

\subsubsection{Pixel losses}

The pixel losses for both hole and non-hole regions target to improve the per-pixel reconstruction accuracy.
\begin{equation}
L_{hole} = \left |  (1 - M) \cdot (I_{out} - I_{gt}) \right |    
\end{equation}
\begin{equation}
L_{valid} = \left |  (M) \cdot (I_{out} - I_{gt}) \right |
\end{equation}

The aggregate loss function is shown in the equation -\ref{tot_loss}. For our initial experiments we use the values of the hyperparameters ($\lambda$s) according to the following equation:
\begin{equation}
\label{tot_loss}
L_{total} = \lambda_{1}L_{tv} + \lambda_{2}L_{valid} + \lambda_{3}L_{hole} + \lambda_{4}L_{perceptual} + \lambda_{5}(L_{style})
\end{equation}

\subsection{Training Details}
The above described architecture, along with the loss functions was trained for 34500 iterations, with a batch size of 3. The totally approximate training time was 20 hours, depending on the GPU system available.

\section{Applications}
We demonstrate the applications of this project in following two different ways.
\subsection{Automated Segmented Object Removal}

\begin{figure}[!h]
\centering
\includegraphics[width=\textwidth, height=\textheight, keepaspectratio]{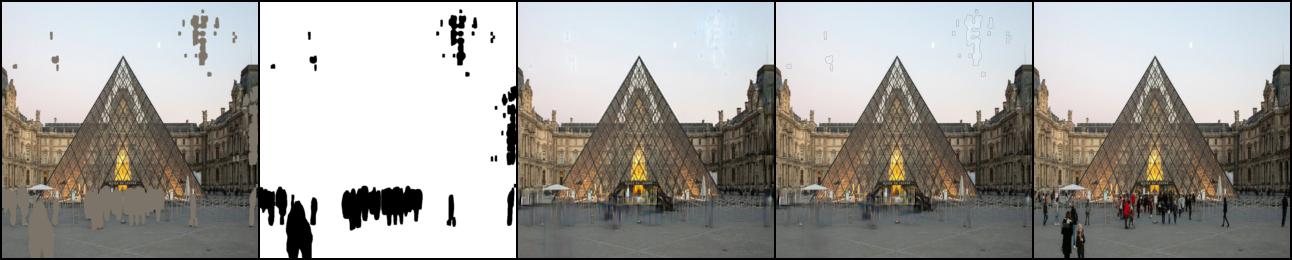}
\caption{Automated Segmented Object Removal, The rightmost image is the ground truth that is provided as input to the automated pipeline. The Image segmentation part of the pipeline detects all the humans and generates a mask for the same as can be seen in the second image from the left. The leftmost image goes as input into the Image Inpainting pipeline which eventually generates reconstruced outputs as seen in the middle and the second last image.}
\label{fig:Louvre}
\end{figure}

In this section, we make use of automatic Image Segmentation Technique with state of the art FCN-ResNet101 Model \cite{shelhamer2016fully}, to generate masks of the segmented regions. The Segmentation model can identify large number of classes (20 in our case), and thus can generate masks corresponding to different objects in the image. We use the pre-trained Pytorch version of this model trained on the Pascal VOC dataset (20 classes) to create this pipeline. For the demonstration of this application, we use the pre-trained Image Inpainting model trained on the Places-2 Dataset. 
The user just needs to give the input image and the automated pipeline firstly detects the humans in the image (any of the 20 classes from the Pascal VOC dataset) and then performs the image inpainting using those detected objects regions as masks. Here we given an example for the same:
In this demonstration, we have chosen the object to be a human figure in the image. The segmentation model creates a Segmented mask as shown in Figure \ref{fig:Louvre}. 

This mask, along with the original image, is fed as an input to our inpainting model. The results are compared with other alternate inpainting algorithms.

\subsection{Manual Mask Generation and Inpainting}
\begin{figure}[!h]
\centering
\includegraphics[width=0.8\textwidth, height=0.8\textheight, keepaspectratio]{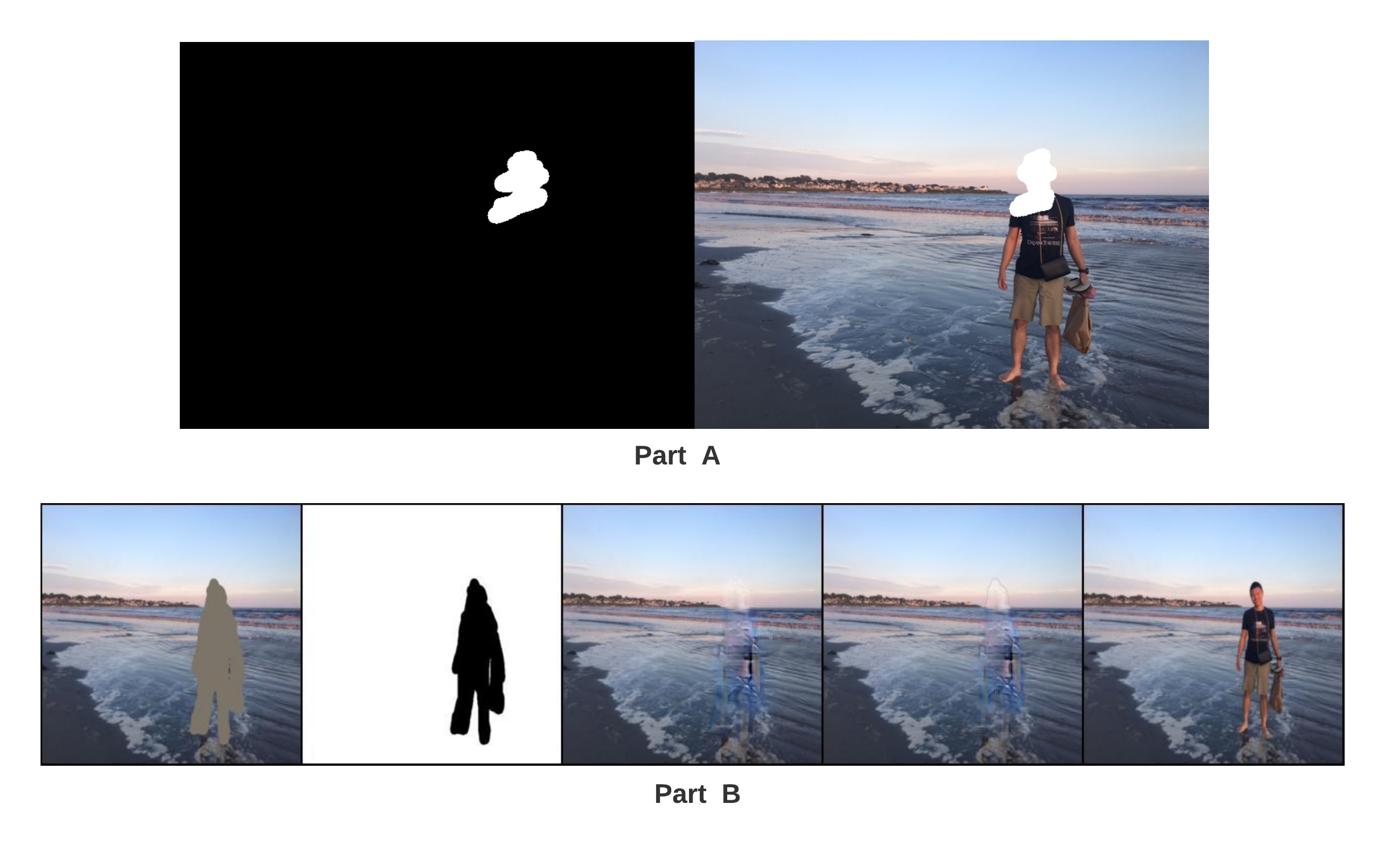}
\caption{Manual Mask Generation and Inpainting. Part A: The user can create a manual mask in the input image. Part B:  The resulting output generated consists of the masked input, the mask, output of our work, output from Navier stokes algorithm, and the ground truth from left to right respectively.}
\label{fig:cmm}
\end{figure}

In this demonstration, we facilitate the user with a custom board where one can create a mask manually. The created mask is then inpainted using our trained model. As with the previous application, this method is also compared with other Classical approaches to compare the results. The results of this experiment are shown in Figure \ref{fig:cmm}.

\section{Observations}

The state of the art models for image feature extraction break the images into primarily two types of information. The first kind of features are the features that are essential for the shape (edges, blobs, corners etc.) of the image, and the other kind of features, are essential for the style (color patterns) of the image. In this project, we tried to understand the contribution of each of these types of features, and study how removing one of them can affect the performance of the architecture in generating the resulting images. Our qualitative and quantitative observations are described in the following subsections.

\subsection{Ablation Study}
An Ablation Study typically involves removing some component of the model, to understand the contribution of the component to the overall system. As described in the previous sections, this work focuses significantly on the complex loss functions. Therefore, we performed the following two experiments. In the first experiment, we removed perceptual loss (described in Section 2.5) component from the loss function of the model, and trained the model for 20000 epochs. In the second experiment, we removed the Style loss component(described in Section 2.5) and similarly trained the model. These experiments are a clear proof of concept for the losses described in Section 2.

\subsubsection{Perceptual Loss Ablation}
Figure 5 shows the results of this experiment. The results of the Perceptual loss ablation clearly reflect the lack of sharp image features, in terms of the edges, corners etc. Although the generated images capture the overall structure of the input image via several other loss functions involved, the results are clearly seem lacking. 

\subsubsection{Style Loss Ablation}
Figure 5 shows the results of this experiment. As described in Section 2.5.3, Style loss is responsible for capturing colors and localised compositions. As we remove the effect of Style loss, we can clearly see that output images contain structural feature details, but are unable to match the ground truth in terms of color compositions.

\begin{figure}[!h]
\centering
\includegraphics[width=0.8\textwidth, height=0.8\textheight, keepaspectratio]{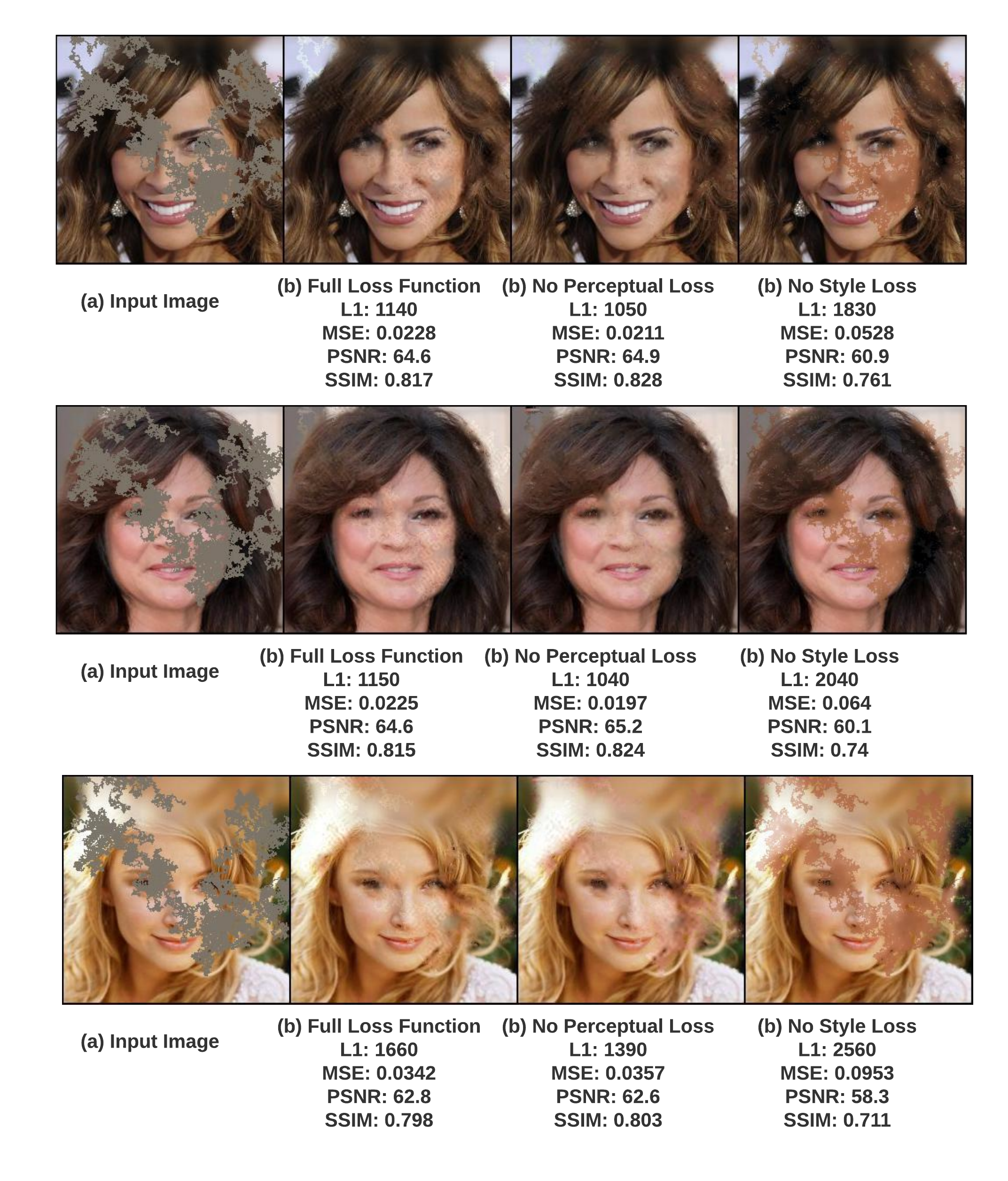}
\caption{Results of Ablation Study UNET}
\label{fig:Unet_arch}
\end{figure}

\subsection{Quantitative Study}
In this section, we make use of the quantitative metric such as L1 Norm, Mean Squared Error(MSE), Peak Signal to Noise Ratio(PSNR), and Structural Similarity Index(SSIM) to measure the performance of our model. The reader must carefully note that there could be multiple possible inpainting results for the same image and mask provided as input. Therefore, although the quantitative analysis gives a measure for comparison different methods, this analysis is not the most robust way for analysing the performance. Previous works in the same field agree on this idea, however, have been reporting these performance metric for comparisons. Therefore, we follow the convention and report our results in terms of the above mentioned metrics.

We compare our outputs with a classical approach named Navier Stokes Image Inpainting Technique. This approach uses ideas from classical fluid dynamics to propagate isophote lines continuously from the exterior into the region to be inpainted. We repeat our study for various mask size to image size ratios. The results are displayed in the Table \ref{tab:loss} below.

\begin{table}[H]
\centering
\setlength\tabcolsep{1pt}
\hspace*{-1cm}
\begin{tabular}
{|
>{\columncolor[HTML]{EFEFEF}}c |
>{\columncolor[HTML]{FFFFFF}}c |
>{\columncolor[HTML]{EFEFEF}}c |
>{\columncolor[HTML]{FFFFFF}}c |
>{\columncolor[HTML]{EFEFEF}}c |
>{\columncolor[HTML]{FFFFFF}}c |
>{\columncolor[HTML]{EFEFEF}}c |
>{\columncolor[HTML]{FFFFFF}}c |
>{\columncolor[HTML]{EFEFEF}}c |
>{\columncolor[HTML]{FFFFFF}}c |
>{\columncolor[HTML]{EFEFEF}}c |
>{\columncolor[HTML]{FFFFFF}}c |
>{\columncolor[HTML]{EFEFEF}}c |}
\hline
\textbf{Trained Model} & \multicolumn{4}{c|}{\cellcolor[HTML]{EFEFEF}\textbf{r=0.05}} & \multicolumn{4}{c|}{\cellcolor[HTML]{EFEFEF}\textbf{r=0.1}} & \multicolumn{4}{c|}{\cellcolor[HTML]{EFEFEF}\textbf{r =0.2}} \\ \hline
 & \cellcolor[HTML]{EFEFEF}l1 & mse & \cellcolor[HTML]{EFEFEF}psnr & \cellcolor[HTML]{EFEFEF}ssim & \cellcolor[HTML]{EFEFEF}l1 & \cellcolor[HTML]{EFEFEF}mse & \cellcolor[HTML]{EFEFEF}psnr & \cellcolor[HTML]{EFEFEF}ssim & \cellcolor[HTML]{EFEFEF}l1 & \cellcolor[HTML]{EFEFEF}mse & \cellcolor[HTML]{EFEFEF}psnr & \cellcolor[HTML]{EFEFEF}ssim \\ \hline
PConv & 
$6.05 \mathrm{e}{+4}$ &
$1.92\mathrm{e}{-1}$ &
$5.55\mathrm{e}{+1}$& 
$4.94\mathrm{e}{-1}$&
$6.81\mathrm{e}{+4}$&
$2.88\mathrm{e}{-1}$&
$5.40\mathrm{e}{+1}$&
$4.66\mathrm{e}{-1}$&
$8.46\mathrm{e}{+4}$&
$9.31\mathrm{e}{-1}$&
$5.09\mathrm{e}{+1}$&
$4.23\mathrm{e}{-1}$\\ \hline
PConv, no Style Loss &
$8.43\mathrm{e}{+4}$&
$3.35\mathrm{e}{-1}$&
$5.33\mathrm{e}{+1}$&
$3.98\mathrm{e}{-1}$&
$1.08\mathrm{e}{+5}$&
$1.43\mathrm{e}{0}$&
$5.02\mathrm{e}{+1}$&
$3.74\mathrm{e}{-1}$&
$2.58\mathrm{e}{+5}$&
$1.29\mathrm{e}{+1}$&
$4.10\mathrm{e}{+1}$&
$3.21\mathrm{e}{-1}$\\ \hline
PConv, no Perceptual Loss &
$1.56\mathrm{e}{+5}$&
$9.83\mathrm{e}{-1}$&
$4.85\mathrm{e}{+1}$&
$1.51\mathrm{e}{-1}$&
$1.70\mathrm{e}{+5}$&
$1.35\mathrm{e}{0}$&
$4.73\mathrm{e}{+1}$&
$1.36\mathrm{e}{-1}$&
$2.29\mathrm{e}{+5}$&
$4.27\mathrm{e}{0}$&
$4.33\mathrm{e}{+1}$&
$1.17\mathrm{e}{-1}$\\ \hline
Classical Method  &
$1.47\mathrm{e}{+7}$&
$1.90\mathrm{e}{+4}$&
$6.34\mathrm{e}{0}$&
$1.43\mathrm{e}{-1}$&
$1.40\mathrm{e}{+7}$&
$1.81\mathrm{e}{+4}$&
$6.55\mathrm{e}{0}$&
$1.35\mathrm{e}{-1}$&
$1.26\mathrm{e}{+7}$&
$1.64\mathrm{e}{+4}$&
$7.04\mathrm{e}{0}$&
$1.20\mathrm{e}{-1}$\\ \hline
\end{tabular}
\caption{Quantitative Measure}
\label{tab:loss}
\end{table}

\subsection{Key Observations}
Some of our key observations are-
\begin{enumerate}
    \item An image is made of structural features and features that correspond to colours.
    \item Classical techniques are not as effective as the learning based approaches for the problem of image inpainting.
    \item Capturing the semantics of the image rather than per pixel information, is a much more efficient method for image reconstruction.
\end{enumerate}

\section{Conclusion and Future Work}

This project was an attempt to reproduce and innovate the Deep Learning based Image Inpainting Methodology by making use of the Partial Convolution Layers and complex Loss Networks. In this project we performed an in-detail analysis of the UNet inspired network architecture, performed ablation studies, and implemented two methodologies to show the application of this technique.

This current project can be extended as a part of other larger Deep learing frameworks such Image Deocclusion, or can be further implemented for automatic modifications to Videos. 





\bibliographystyle{IEEEtran}
\bibliography{refs}

\end{document}